%% file: 00-main.tex
\documentclass[11pt,a4paper]{article}
\usepackage[hyperref]{eacl2021}
\usepackage{times}
\usepackage{latexsym}

\usepackage[english]{babel}
\usepackage{blindtext}
\usepackage{microtype}

\aclfinalcopy 

\usepackage{graphicx}
\usepackage{amsthm}
\usepackage{amsmath}
\usepackage{booktabs}
\usepackage{algorithm}
\usepackage{algorithmic}
\usepackage{multirow}
\usepackage{subcaption}
\usepackage{mathtools}
\usepackage{amssymb}
\usepackage{tabularx}
\usepackage{adjustbox}
\usepackage{arydshln}
\usepackage{tikz}
\usetikzlibrary{bayesnet}

\definecolor{esc}{RGB}{0, 153, 51}

\definecolor{yingzhen}{RGB}{0, 153, 51}
\definecolor{ehsan}{RGB}{0, 40, 255}
\definecolor{yi}{RGB}{255, 140, 0}

\title{Combining Deep Generative Models and Multi-lingual Pretraining\\ for Semi-supervised Document Classification} 

\author{
  Yi Zhu$^\heartsuit$
  \ \ Ehsan Shareghi$^\spadesuit$$^\heartsuit$
  \ \ Yingzhen Li$^\diamondsuit$\thanks{\enspace Work done while at Microsoft Research Cambridge.}
  \ \ \ Roi Reichart$^\clubsuit$
  \ \ Anna Korhonen$^\heartsuit$\\
  $^\heartsuit$~Language Technology Lab, University of Cambridge\\
  $^\spadesuit$~Department of Data Science \& AI, Monash University \\
  $^\diamondsuit$~Department of Computing, Imperial College London\\
  $^\clubsuit$~Faculty of Industrial Engineering and Management, Technion, IIT\\
  {\tt \{yz568,alk23\}@cam.ac.uk}, {\tt ehsan.shareghi@monash.edu}\\
  {\tt yingzhen.li@imperial.ac.uk}, {\tt roiri@technion.ac.il}
}

\date{}

\begin{document}
\maketitle

\begin{abstract}
\input{01-abstract.tex}
\end{abstract}

\input{02-intro.tex}
\input{03-background.tex}
\input{04-model.tex}
\input{05-experiment.tex}
\input{06-conclusion.tex}

\section*{Acknowledgments}
This work is supported by the ERC Consolidator Grant LEXICAL: Lexical Acquisition Across Languages (648909). 
The first author would like to thank Victor Prokhorov and Xiaoyu Shen for their comments on this work.
The authors would like to thank the three anonymous reviewers for their helpful suggestions.

\bibliography{local}
\bibliographystyle{acl_natbib}

\appendix
\input{appendix.tex}

\end{document}

%% file: 01-abstract.tex
Semi-supervised learning through deep generative models and multi-lingual pretraining techniques have orchestrated tremendous success across different areas of NLP. Nonetheless, their development has happened in isolation, while the combination of both could potentially be effective for tackling task-specific labelled data shortage. To bridge this gap, we combine semi-supervised deep generative models and multi-lingual pretraining to form a pipeline for document classification task. Compared to strong supervised learning baselines, our semi-supervised classification framework is highly competitive and outperforms the state-of-the-art counterparts in low-resource settings across several languages.~\footnote{Code is available at \url{https://github.com/cambridgeltl/mling_sdgms}.}

%% file: 02-intro.tex
\section{Introduction}

Multi-lingual pretraining has been shown to effectively use unlabelled data through learning shared representations across languages that can be transferred to downstream tasks~\cite{DBLP:journals/tacl/ArtetxeS19,DBLP:conf/naacl/DevlinCLT19,DBLP:conf/emnlp/WuD19,DBLP:conf/nips/ConneauL19}. Nonetheless, the lack of labelled data still leads to inferior performance of the same model compared to those trained in languages with more labelled data such as English~ \cite{DBLP:conf/conll/ZemanHPPSGNP18,zhu-etal-2019-importance}. 

Semi-supervised learning is another appealing paradigm that supplements the labelled data with unlabelled data which is easy to acquire~\cite[\textit{inter alia}]{DBLP:conf/colt/BlumM98,DBLP:journals/tkde/ZhouL05,DBLP:conf/naacl/McCloskyCJ06}. In particular, deep generative models (DGMs) such as variational autoencoder (VAE; \citet{DBLP:journals/corr/KingmaW13}) are capable of capturing complex data distributions at scale with rich latent representations, and they
have been used for semi-supervised learning in various tasks in NLP~\cite{DBLP:conf/aaai/XuSDT17,DBLP:conf/acl/NeubigZYH18,DBLP:conf/acl/ChoiKL19,ijcai2019-737}, as well as inducing cross-lingual word embeddings~\cite{ijcai2017-582}, and representation learning in combination with Transformers via pretraining \citep{DBLP:journals/corr/abs-2004-04092}.


To leverage the benefits of both worlds, we propose a pipeline method by combining semi-supervised DGMs (SDGMs) based on M1+M2 model 
~\cite{NIPS2014_5352}
with multi-lingual pretraining. The pretrained model serves as multi-lingual encoder, and SDGMs can operate on top of it independently of encoding architecture. To highlight such independence, we experiment with two pretraining settings: (1) our LSTM-based cross-lingual VAE, and (2) the current stat-of-the-art (SOTA) multi-lingual BERT \citep{DBLP:conf/naacl/DevlinCLT19}.

Our experiments on document classification in several languages show promising
results via the SDGM framework with different encoders, outperforming the SOTA supervised counterparts. We also illustrate that the end-to-end training of M1+M2 that was previously considered too unstable to train~\cite{DBLP:conf/icml/MaaloeSSW16} is possible with a reformulation of the objective function.

\begin{table*}[t]
    \centering
    \resizebox{\linewidth}{!}{
    \begin{tabular}{lc}
    $\mathcal{L}(\mathbf{x}, y) = \underbrace{
\mathbb{E}_{q_{\phi}(\mathbf{z}_1 | \mathbf{x})}
[
\log p_{\theta}(\mathbf{x} | \mathbf{z}_1)
]
}_\text{Reconstruction}
-
\underbrace{
\mathbb{E}_{q_{\phi}(\mathbf{z}_1 | \mathbf{x})q_{\phi}(\mathbf{z}_2 | \mathbf{z}_1, y)}
[
\log \frac{q_{\phi}(\mathbf{z}_2 | \mathbf{z}_1, y)}{p(\mathbf{z}_2)} 
+ 
\log \frac{q_{\phi}(\mathbf{z}_1 | \mathbf{x})}{p_{\theta}(\mathbf{z}_1 | \mathbf{z}_2, y)} 
]
}_\text{KL}
+
\underbrace{
\log p(y)
}_\text{Constant}
$&\multirow{2}[2]{*}[2mm]{  \includegraphics[scale=1.3]{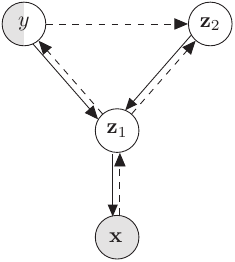}}\\
&\\
         $\mathcal{U}(\mathbf{x})=\underbrace{
\mathbb{E}_{q_{\phi}(\mathbf{z}_1 | \mathbf{x})}
[\log p_{\theta}(\mathbf{x} | \mathbf{z}_1)]
}_\text{Reconstruction}
-
\underbrace{
\mathbb{E}_{q_{\phi}(\mathbf{z}_1 | \mathbf{x})q_{\phi}(y | \mathbf{z}_1)q_{\phi}(\mathbf{z}_2 | \mathbf{z}_1, y)}
[\log \frac{q_{\phi}(\mathbf{z}_2 | \mathbf{z}_1, y)}{p(\mathbf{z}_2)} + 
\log \frac{q_{\phi}(\mathbf{z}_1 | \mathbf{x})}{p_{\theta}(\mathbf{z}_1 | \mathbf{z}_2, y)} +
\log \frac{q_{\phi}(y | \mathbf{z}_1)}{p(y)}
]
}_\text{KL}$&\\
    \end{tabular}}
    \vspace{-2mm}
    \caption{Labelled and unlabelled objectives for M1+M2 model (left), and its corresponding graphical model (right).}
    \label{tab:equations}
    \vspace{-2mm}
\end{table*}

%% file: 03-background.tex
\section{Semi-supervised Learning with DGMs}
\paragraph{Variational Autoencoder.}
VAE consists of a stochastic neural encoder $q_{\phi}(\mathbf{z}|\mathbf{x})$ that maps an input $\mathbf{x}$ to a latent representation $\mathbf{z}$, and a neural decoder $p_{\theta}(\mathbf{x}| \mathbf{z})$ that reconstructs $\mathbf{x}$, jointly trained by maximising the evidence lower bound (ELBO) of the marginal likelihood of the data:
\begin{equation}\label{eq:vae}
\setlength{\abovedisplayskip}{3pt}
\setlength{\belowdisplayskip}{3pt}
\setlength{\abovedisplayshortskip}{3pt}
\setlength{\belowdisplayshortskip}{3pt}
\resizebox{0.89\linewidth}{!}{
  \begin{minipage}{\linewidth}
\centering
$
\begin{aligned}
\mathbb{E}_{q_{\phi}(\mathbf{z} | \mathbf{x})}\big[\log p_{\theta}(\mathbf{x} | \mathbf{z})\big] - \textrm{KL}\big(q_{\phi}(\mathbf{z} | \mathbf{x}) \| p(\mathbf{z})\big)
\end{aligned}
$
\end{minipage}}
\end{equation}
where the first term (reconstruction) maximises the expectation of data likelihood under the posterior distribution of $\mathbf{z}$, and the Kullback-Leibler (KL) divergence regulates the distance between the learned posterior and prior of $\mathbf{z}$.

\paragraph{Semi-supervised Learning with VAEs.}
The SDGM we use for semi-supervised learning is M1+M2 \citep{NIPS2014_5352}, a graphical model (Table \ref{tab:equations}~(right)), with two layers of stochastic variables $\mathbf{z}_1$ and $\mathbf{z}_2$, with each being an isotropic Gaussian distribution. 
The first layer encodes the input sequence $\mathbf{x}$ into a deterministic hidden representation $\mathbf{h}$, and outputs the posterior distribution of $\mathbf{z}_1$:
\begin{equation}\label{eq:vae_infer}
\setlength{\abovedisplayskip}{-10pt}
\setlength{\belowdisplayskip}{3pt}
\setlength{\abovedisplayshortskip}{3pt}
\setlength{\belowdisplayshortskip}{3pt}
\resizebox{0.89\linewidth}{!}{
  \begin{minipage}{\linewidth}
\centering
$
\begin{aligned}
q_{\phi}(\mathbf{z}_1 | \mathbf{x}) = \mathcal{N}\Big(\boldsymbol{\mu}_{\phi}(\mathbf{h}), \textrm{diag}\big(\boldsymbol{\sigma}_{\phi}^2(\mathbf{h})\big)\Big)
\end{aligned}
$
\end{minipage}}
\end{equation}
As our SDGM is independent of the encoding architecture, we use different pretrained multi-lingual models to obtain $\mathbf{h}$, $\boldsymbol{\mu}_{\phi}(\mathbf{h})$, and $\boldsymbol{\sigma}_{\phi}^2(\mathbf{h})$, described in \S\ref{sec:nxvae}.
The second layer computes the posterior distribution of $\mathbf{z}_2$, conditioned on sampled $\mathbf{z}_1$ from $q_{\phi}(\mathbf{z}_1 | \mathbf{x})$ and a class variable $y$.

When we use labelled data, i.e. $y$ is observed, $q_{\phi}(\mathbf{z}_2 | \mathbf{z}_1, y)$ can be directly obtained.
With unlabelled data, we calculate the posterior $q_{\phi}(\mathbf{z}_2, y | \mathbf{z}_1) = q_{\phi}(y | \mathbf{z}_1)q_{\phi}(\mathbf{z}_2 | \mathbf{z}_1, y)$ by inferring $y$ with the classifier $q_{\phi}(y | \mathbf{z}_1)$, and integrate over all possible values of $y$.
%
%
Therefore, the ELBO for the labelled data $\mathcal{S}_l = \{\mathbf{x}, y\}$ is $\mathcal{L}(\mathbf{x}, y)$:
\begin{equation*}
\setlength{\abovedisplayskip}{3pt}
\setlength{\belowdisplayskip}{3pt}
\setlength{\abovedisplayshortskip}{3pt}
\setlength{\belowdisplayshortskip}{3pt}
\resizebox{0.95\linewidth}{!}{
  \begin{minipage}{\linewidth}
\centering
$
\begin{aligned}
\mathbb{E}_{q_{\phi}(\mathbf{z}_1, \mathbf{z}_2 | \mathbf{x}, y)}\Big[\!\log\frac{p_{\theta}(\mathbf{x}, y, \mathbf{z}_1, \mathbf{z}_2)}{q_{\phi}(\mathbf{z}_1, \mathbf{z}_2 | \mathbf{x}, y)}\Big]\! \le\! \log p(\mathbf{x}, y)
\end{aligned}
$
\end{minipage}}
\end{equation*}
and for the unlabelled data $\mathcal{S}_u = \{\mathbf{x}\}$ is $\mathcal{U}(\mathbf{x})$:
\begin{equation*}
\setlength{\abovedisplayskip}{3pt}
\setlength{\belowdisplayskip}{3pt}
\setlength{\abovedisplayshortskip}{3pt}
\setlength{\belowdisplayshortskip}{3pt}
\resizebox{0.95\linewidth}{!}{
  \begin{minipage}{\linewidth}
\centering
$
\begin{aligned}
\mathbb{E}_{q_{\phi}(\mathbf{z}_1, \mathbf{z}_2, y | \mathbf{x})}\Big[\!\log\frac{p_{\theta}(\mathbf{x}, y, \mathbf{z}_1, \mathbf{z}_2)}{q_{\phi}(\mathbf{z}_1, \mathbf{z}_2, y | \mathbf{x})}\Big]\! \le\! \log p(\mathbf{x}) 
\end{aligned}
$
\end{minipage}}
\end{equation*}
where the generation part is $p_{\theta}(\mathbf{x}, y, \mathbf{z}_1, \mathbf{z}_2)=p(y)p(\mathbf{z}_2)p_{\theta}(\mathbf{z}_1|\mathbf{z}_2, y)p_{\theta}(\mathbf{x}|\mathbf{z}_1)$, $p(y)$ is uniform distribution as the prior of $y$, $p(\mathbf{z}_2)$ is standard Gaussian distribution as the prior of $\mathbf{z}_2$, and $p_{\theta}(\mathbf{x}|\mathbf{z}_1)$ is the decoder, which can have different architectures depending on the encoder (\S\ref{sec:experiment}).

The objective function maximises both the labelled and unlabelled ELBOs while training directly the classifier with the labelled data as well:
\begin{equation*}
\setlength{\abovedisplayskip}{3pt}
\setlength{\belowdisplayskip}{3pt}
\setlength{\abovedisplayshortskip}{3pt}
\setlength{\belowdisplayshortskip}{3pt}
\resizebox{0.95\linewidth}{!}{
  \begin{minipage}{\linewidth}
$
\begin{aligned}
\mathcal{J} =\!\!
\sum_{(\mathbf{x}, y) \in \mathcal{S}_l} \!\!\big(\mathcal{L}(\mathbf{x}, y) + \alpha \mathcal{J}_{cls}(\mathbf{x}, y)\big) + \!\!\sum_{\mathbf{x} \in \mathcal{S}_u} \mathcal{U}(\mathbf{x})
\end{aligned}
$
\end{minipage}}
\end{equation*}
where $\mathcal{J}_{cls}(\mathbf{x}, y) = \mathbb{E}_{q_{\phi}(\mathbf{z}_1 | \mathbf{x})}[q_{\phi}(y | \mathbf{z}_1)]$, 
and $\alpha$ is a hyperparameter to tune. 
Considering the factorisation of the model according to the graphical model, we can rewrite the $\mathcal{L}(\mathbf{x}, y)$ and $\mathcal{U}(\mathbf{x})$ as shown in Table~\ref{tab:equations}(left).
The reconstruction term is the expected log likelihood of the input sequence $\mathbf{x}$, same for both ELBOs.
The KL term regularises the posterior distributions of $\mathbf{z}_1$ and $\mathbf{z}_2$ according to their priors.
Additionally for $\mathcal{U}(\mathbf{x})$, as mentioned before, we first infer $y$ and treat it as if it were observed, so we need to compute the expected KL term over $q_{\phi}(y | \mathbf{z}_1)$ regularised by $\textrm{KL}(q_{\phi}(y | \mathbf{z}_1) \| p(y))$.


Due to its training difficulty, M1+M2 is trained layer-wise in \citet{NIPS2014_5352}, where the first layer is trained according to Eq.~\ref{eq:vae} and fixed, before the second layer is trained on top. However, in our experiments (\S\ref{sec:mono}) we found that M1+M2 is easier to train end-to-end.
We attribute this to our mathematical reformulation of the objective functions, giving rise to a more stable optimisation schedule.


%% file: 04-model.tex
\section{SDGMs with Multi-lingual Pretraining}\label{sec:nxvae}
\paragraph{LSTM-based Encoder with VAE Pretraining.}
Our pretraining is based on the framework of \citet{ijcai2017-582}, in which they pretrain a cross-lingual VAE with parallel corpus as input. However, the parallel corpus is expensive to obtain, and only the resulting cross-lingual embeddings rather than the whole encoder could be used due to the parallel input limitation of the model. 
To address these shortcomings, we propose non-parallel cross-lingual VAE (NXVAE),
which has the same graphical model as the vanilla VAE. 
Each language $i$ is associated with its own word embedding matrix, and its input sequence $\mathbf{x}_i$ is processed via a two layer BiLSTM \citep{Hochreiter:1997:LSM:1246443.1246450} shared across languages.
We use the concatenation of the BiLSTM last hidden states as $\mathbf{h}$, and compute $q_{\phi}(\mathbf{z} | \mathbf{x}_{i})$ with Eq. \ref{eq:vae_infer}, so that $\mathbf{z}$ becomes the joint cross-lingual semantic space. 
A language specific bag-of-word decoder (BOW; \citet{pmlr-v48-miao16}) is then used to reconstruct the input sequence.
Additionally, we optimise a language discriminator as an adversary~\cite{lample2018unsupervised} to encourage the mixing of different language representations and keep the shared encoder language-agnostic. 
After pretraining NXVAE, we transfer the whole encoder, including $\boldsymbol{\mu}_{\phi}(\mathbf{h})$ and $\boldsymbol{\sigma}_{\phi}^2(\mathbf{h})$, directly into our SDGM framework and treat it as $q_{\phi}(\mathbf{z}_1 | \mathbf{x})$ component of the model (\S\ref{sec:mono}).

\paragraph{Multi-lingual BERT Encoder.} 
To show that our SDGM is effective with other encoding architectures, we use the pretrained multi-lingual BERT (mBERT; \citet{DBLP:conf/naacl/DevlinCLT19})\footnote{\url{https://github.com/google-research/bert/blob/master/multilingual.md}.} as our encoder.
Given an input sequence, the pooled \texttt{[CLS]} representation is used as $\mathbf{h}$ to compute $q_{\phi}(\mathbf{z}_1 | \mathbf{x})$~(Eq.~\ref{eq:vae_infer}).
Different from NXVAE, we initialise the parameters of $\boldsymbol{\mu}_{\phi}(\mathbf{h})$ and $\boldsymbol{\sigma}_{\phi}^2(\mathbf{h})$ randomly.

%% file: 05-experiment.tex
\section{Experiments}\label{sec:experiment}
We perform document classification on the class balanced multilingual document classification corpus (MLDoc; \citet{DBLP:conf/lrec/SchwenkL18}).
Each document is assigned to one of the four news topic classes: \textit{corporate/industrial} (C), \textit{economics} (E), \textit{government/social} (G), and \textit{markets} (M).
We experiment with five representative languages: \textsc{en}, \textsc{de}, \textsc{fr}, \textsc{ru}, \textsc{zh}, and use $1\textrm{k}$ instance training set along with the standard development and test set.
For experiments with varying labelled data size, the rest training data from $1\textrm{k}$ corpus is used as unlabelled data.
The full statistics are shown in Table~\ref{tb:mldoc_info}.
Three languages (\textsc{en}, \textsc{de}, \textsc{fr}) are tested for LSTM encoder with VAE pretraining (\S\ref{sec:mono}) and all five languages for mBERT encoder (\S\ref{sec:bert}).
All documents are lowercased.
We report \textit{accuracy} for evaluation following~\citet{DBLP:conf/lrec/SchwenkL18}. 

For all experiments, We use Adam~\citep{DBLP:journals/corr/KingmaB14} as optimiser, but with different learning rates for both settings and pretraining.
We implemented the model with Pytorch\footnote{\url{https://pytorch.org/}.} 1.10~\citep{paszke2019pytorch}, and use GeForce GTX 1080Ti GPUs.
See the Appendix for details about model configurations
and training.

\subsection{LSTM Encoder with VAE Pretraining}\label{sec:mono}
\paragraph{Experimental Setup.} For pretraining NXVAE, we use three language pairs: \textsc{en-de}, \textsc{en-fr} and \textsc{de-fr} constructed from Europarl v7 parallel corpus \citep{Koehn2002EuroparlAM},\footnote{\url{https://www.statmt.org/europarl/}.} where only two language pairs are available:
\textsc{en}-\textsc{de} and \textsc{en}-\textsc{fr}, which consist of four datasets in total: (\textsc{en}, \textsc{de})$_{\textrm{\textsc{en}-\textsc{de}}}$, and (\textsc{en}, \textsc{fr})$_{\textrm{\textsc{en}-\textsc{fr}}}$.
For \textsc{de}-\textsc{fr}, we pair \textsc{de}$_{\textrm{\textsc{en}-\textsc{de}}}$ and \textsc{fr}$_{\textrm{\textsc{en}-\textsc{fr}}}$ directly as pseudo parallel data. 
We trim all datasets into exactly the same sentence size, and preprocess them with: tokenization, lowercasing, substituting digits with $0$, and removing all punctuations, redundant spaces and empty lines.
We randomly sample a small part of parallel sentences to build a development set.
For models which do not require parallel input, e.g. NXVAE, we mix the two datasets of a language pair together.
To avoid KL-collapse during pretraining, a weight $\alpha$ on the KL term in Eq.~\ref{eq:vae} is tuned and fixed to $0.1$~\citep{DBLP:conf/iclr/HigginsMPBGBML17,DBLP:conf/icml/AlemiPFDS018}. 
We only run one trial with fixed random seed for both pretraining and document classification. 
Training details can be found in the Appendix.

\begin{table}[t]
	\centering
    \def\arraystretch{0.95}
    {\footnotesize
	\begin{tabularx}{\columnwidth}{l XXXXX}
	\toprule
		
		\multicolumn{1}{l}{} & \textsc{C} & \textsc{E} & \textsc{G} & \textsc{M} & Total\\ \toprule
		
		\multirow{3}{*}{\textsc{en}} &  
	    270 & 234 & 252 & 244 & 1000\\
	    & 228 & 238 & 266 & 268 & 1000\\
	    & 991 & 1000 & 1030 & 979 & 4000\\\cmidrule(lr){2-6}
	    
		\multirow{3}{*}{\textsc{de}} &  
		270 & 240 & 245 & 245 & 1000\\
	    & 229 & 268 & 266 & 237 & 1000\\
	    & 984 & 1026 & 1022 & 968 & 4000\\\cmidrule(lr){2-6}
	    
		\multirow{3}{*}{\textsc{fr}} &  
		227 & 262 & 258 & 253 & 1000\\
		& 257 & 237 & 237 & 269 & 1000\\
		& 999 & 973 & 998 & 1030 & 4000\\\cmidrule(lr){2-6}
		
		\multirow{3}{*}{\textsc{ru}} &  
		261 & 288 & 184 & 267 & 1000\\
		& 265 & 272 & 204 & 259 & 100\\
		& 1073 & 1121 & 706 & 1100 & 4000\\\cmidrule(lr){2-6}
		
		\multirow{3}{*}{\textsc{zh}} &  
		294 & 286 & 109 & 311 & 1000\\
		& 324 & 300 & 93 & 283 & 1000\\
		& 1169 & 1215 & 363 & 1253 & 4000\\
		
		\bottomrule       
	\end{tabularx}}%
    \vspace{-2mm}
     \caption{Statistics of MLDoc in five languages.
     Instance numbers for each class along with the total numbers are shown. 
     For each language, three rows are training, development and test set instance numbers.} \label{tb:mldoc_info}
     \vspace{-2mm}
\end{table}

As our supervised baselines we compare with the following two groups:
(I) NXVAE-based supervised models which are pretrained NXVAE encoder with a multi-layer perceptron classifier on top (denoted by NXVAE-z$_1$ ($q_{\phi}(y | \mathbf{z}_1)$) or NXVAE-h ($q_{\phi}(y | \mathbf{h})$) depending on the representation fed into the classifier; or 
NXVAE-z$_1$ models initialised with different pretrained embeddings: random initialisation (RAND), mono-lingual fastText (FT; \citet{Q17-1010}), unsupervised cross-lingual MUSE \cite{lample2018word}, pretrained embeddings from \citet{ijcai2017-582} (PEMB), and our resulting embeddings from pretrained NXVAE (NXEMB).\footnote{All embeddings are pretrained on the same Europarl data.} 
(II) We also pretrain a word-based BERT~(BERTW) with parameter size akin to NXVAE on the same data, and fine-tune it directly.\footnote{
We also trained subword-based models for BERT and NXVAE, and observed similar trends. See the Appendix.}

For our semi-supervised experiments, we test two types of decoders with different model capacities: BOW and GRU \cite{cho-al-emnlp14}.
We use M1+M2+BOW (GRU) to denote the model with joint training using a specific decoder, and M1+M2 to denote the original model in \citet{NIPS2014_5352} with layer-wise training.\footnote{We also compared this against a more complex Skip Deep Generative Model \citep{DBLP:conf/icml/MaaloeSSW16}, but found that end-to-end M1+M2 performs better. Details in the Appendix.} We also add a semi-supervised self-training method~\citep{DBLP:conf/naacl/McCloskyCJ06} for BERTW to leverage the unlabelled data (BERTW+ST), where we iteratively add predicted unlabelled data when the model achieves a better dev. accuracy until convergence.

\begin{table}[t]
	\centering
	{\footnotesize
    \setlength{\tabcolsep}{4.5pt}
	\begin{tabularx}{\columnwidth}{ll X}
	\toprule
		Word pair & Lang & kNNs ($k = 3$)\\\toprule
		                                       
		\multirow{2}{*}{president (\textsc{en})}    & \textsc{en} & mr, madam, gentlemen\\
		                                            & \textsc{de} & pr\"asident, herr, kommissar\\
        \addlinespace[0.75ex]
		\multirow{2}{*}{pr\"asident (\textsc{de})}  & \textsc{en} & president, mr, madam\\
		                                            & \textsc{de} & herr, kommissar, herren\\
		                                            \cmidrule(lr){1-3}
		                                            
		\multirow{2}{*}{great (\textsc{en})}        & \textsc{en} & deal, with, a\\
		                                            & \textsc{de} & gro{\ss}e, eine, gute\\
        \addlinespace[0.75ex]
		\multirow{2}{*}{gro{\ss} (\textsc{de})}     & \textsc{en} & striking, gets, lucrative\\
                                                    & \textsc{de} & gering, heikel, hoch\\
                              \cmidrule(lr){1-3}
		                                            
		\multirow{2}{*}{said (\textsc{en})}         & \textsc{en} & already, as, been\\
		                                            & \textsc{de} & gesagt, mit, dem\\
        \addlinespace[0.75ex]
		\multirow{2}{*}{sagte (\textsc{de})}        & \textsc{en} & he, rightly, said\\
		                                            & \textsc{de} & vorhin, kollege, kommissar\\
		\bottomrule 
	\end{tabularx}}
    \vspace{-2mm}
    \caption{Cosine similarity-based nearest neighbours of words (left column) in embedding spaces of \textsc{en} and \textsc{de}. 
    } \label{tb:emb}
    \vspace{-2mm}
\end{table}

\paragraph{Qualitative Results.} 
Table \ref{tb:emb} illustrates the quality of the learned alignments in the cross-lingual space of NXVAE for \textsc{en-de} word pairs.
\paragraph{Classification Results.} 
Table~\ref{tb:semimono} (\textbf{\textsc{en-de}}) shows that within supervised models the NXVAE-z$_1$ substantially outperforms other supervised baselines with the exception of BERTW. The fact that NXVAE-z$_1$ is significantly better than NXVAE-h, suggests that pretraining has enabled $\mathbf{z}_1$ to learn more general knowledge transferable to this task.
Combining with SDGMs, our best pipeline outperforms all baselines across data sizes and languages, including BERTW+ST with bigger gaps in fewer labelled data scenario. We observe the same trend of performance in both supervised and semi-supervised DGM settings on \textbf{\textsc{en-fr}} and \textbf{\textsc{de-fr}}.

For decoder, BOW outperforms the GRU, a finding in line with the results of \citet{DBLP:journals/corr/abs-1910-11856} which suggests
a few keywords seem to suffice for this task.
The poor performance of the original M1+M2, implies the domain discrepancy between pretraining and task data, and highlights the impact of fine-tuning.
In addition, our NXEMB, as a byproduct of NXVAE, performs comparably well with MUSE, and better than all other embedding models including its parallel counterpart PEMB. 

\begin{table}[t]
	\centering
    \footnotesize
    \setlength{\tabcolsep}{2pt}
	\begin{tabularx}{\columnwidth}{@{}l cccc cccc}
	\toprule
\# Labels& \textsc{32} & \textsc{64} & \textsc{128} & \textsc{1k} &
		\textsc{32} & \textsc{64} & \textsc{128} & \textsc{1k}\\ \toprule 
	    \textbf{\textsc{en-de}} & \multicolumn{4}{c}{\textsc{en}} & \multicolumn{4}{c}{\textsc{de}}\\ 
	    \cmidrule[1.5pt](lr){2-5}\cmidrule[1.5pt](lr){6-9}
		
		\textsc{nxvae}-h    & 56.5          & 61.7          & 59.5          & 78.4          & 53.6          & 66.7          & 78.9          & 87.2\\
	    \textsc{nxvae}-z$_1$    & 63.9          & 71.4          & 77.0          & 91.6          & \textbf{65.0} & 73.8          & 82.7          & 93.0\\
	    \textsc{rand}       & 50.1          & 54.2          & 62.3          & 82.5          & 47.2          & 60.8          & 69.0          & 84.8\\     
	    \textsc{ft}         & 36.3          & 49.4          & 61.1          & 80.9          & 45.0          & 54.3          & 69.2          & 86.1 \\
	    \textsc{muse}       & 59.8          &  65.4 & 71.8 & 88.4 & 45.1          & 66.2          & 79.7 & 90.4\\
	    \textsc{pemb}      & 36.4          & 53.9          & 50.9          & 84.4          & 39.4          & 52.0          & 69.0          & 86.7\\
	    \textsc{nxemb}      & 61.5 & 62.0          & 68.6          & 85.4          & 53.4 & 71.2 & 75.9          & 88.8\\
	    \cdashline{2-9}
	    	    \textsc{bertw}      & \textbf{67.7} & \textbf{72.7} & \textbf{84.6} & \textbf{91.8} & 58.1          & \textbf{77.5} & \textbf{89.2} & \textbf{94.0}\\ 
	    \cmidrule(lr){2-5}\cmidrule(lr){6-9}

	    \textsc{m1+m2}      & 56.6          & 67.1          & 70.3          & -    & 52.6           & 67.2          & 76.8          & -\\
	    \textsc{m1+m2+bow}  & \textbf{79.8} & \textbf{81.7} & \textbf{87.2}          & -    & 70.5           & 79.6          & \textbf{89.7} & -\\ 
	    \textsc{m1+m2+gru}  & 75.3          & 79.4          & 84.9          & -    & \textbf{75.1}           & \textbf{80.0}          & 87.1          & -\\
	    \cdashline{2-9}
	    \textsc{bertw+st}  & 68.4 & 73.9 & 86.4 & - & 59.6 & 79.7 & 89.4 & -\\ 
	\bottomrule
	
	    \textbf{\textsc{en-fr}}& \multicolumn{4}{c}{\textsc{en}} & \multicolumn{4}{c}{\textsc{fr}}\\ 
	    \cmidrule[1.5pt](lr){2-5}\cmidrule[1.5pt](lr){6-9}
		\textsc{nxvae}-h        & \textbf{71.4} & 73.8          & 78.6          & 88.0          & 62.8          & 72.7          & 79.9          & 88.9\\
	    \textsc{nxvae}-z$_1$    & 71.2          & \textbf{75.3} & \textbf{80.4} & \textbf{91.2} & \textbf{68.3} & \textbf{75.0} & \textbf{81.4} & \textbf{91.7}\\
	    \cmidrule(lr){2-5}\cmidrule(lr){6-9}
	    \textsc{m1+m2}      & 71.8          & 73.5          & 76.5          & -    & 66.2           & 78.7          & 79.7          & -\\
	    \textsc{m1+m2+bow}  & \textbf{81.0} & \textbf{85.5} & \textbf{88.2} & -    & 80.3           & \textbf{86.0} & \textbf{88.8} & -\\ 
	    \textsc{m1+m2+gru}  & 75.3          & 81.4          & 83.8          & -    & \textbf{80.7}  & 82.3          & 87.4          & -\\ 
	   \toprule
	   \textbf{\textsc{de-fr}} & \multicolumn{4}{c}{\textsc{de}} & \multicolumn{4}{c}{\textsc{fr}}\\ 
	    \cmidrule[1.5pt](lr){2-5}\cmidrule[1.5pt](lr){6-9}
		\textsc{nxvae}-h        & 42.4          & 53.3          & 74.3          & 85.7          &  39.8   & 51.8      &    58.5    & 86.9\\
	    \textsc{nxvae}-z$_1$    & \textbf{63.3} & \textbf{75.4} & \textbf{81.3} & \textbf{92.1} & \textbf{60.1} & \textbf{71.1} & \textbf{78.4} & \textbf{91.4}\\
	    \cmidrule(lr){2-5}\cmidrule(lr){6-9}
	    \textsc{m1+m2}      & 59.1          & 70.6          & 75.4          & -    & 48.5           & 57.4          & 60.7          & -\\ 
	    \textsc{m1+m2+bow}  & \textbf{78.0} & \textbf{83.2} & \textbf{88.3}          & -    & \textbf{81.4}  & \textbf{84.5} & \textbf{88.4} & -\\ 
	    \textsc{m1+m2+gru}  & 74.6          & 80.5          & 86.2         & -    & 66.2           & 77.2          & 81.9          & -\\ 
	\bottomrule
	\end{tabularx}
    \vspace{-2mm}
     \caption{MLDoc test accuracy for \textsc{en-de}, \textsc{en-fr} and \textsc{de-fr} pairs. 
     The best results for supervised and semi-supervised models are in bold.} \label{tb:semimono}
     \vspace{-2mm}
\end{table}

\begin{figure*}[t]
	\centering
	\includegraphics[width=0.97\linewidth, trim={0.2cm 0.9cm 0.2cm 1.4cm}, clip]{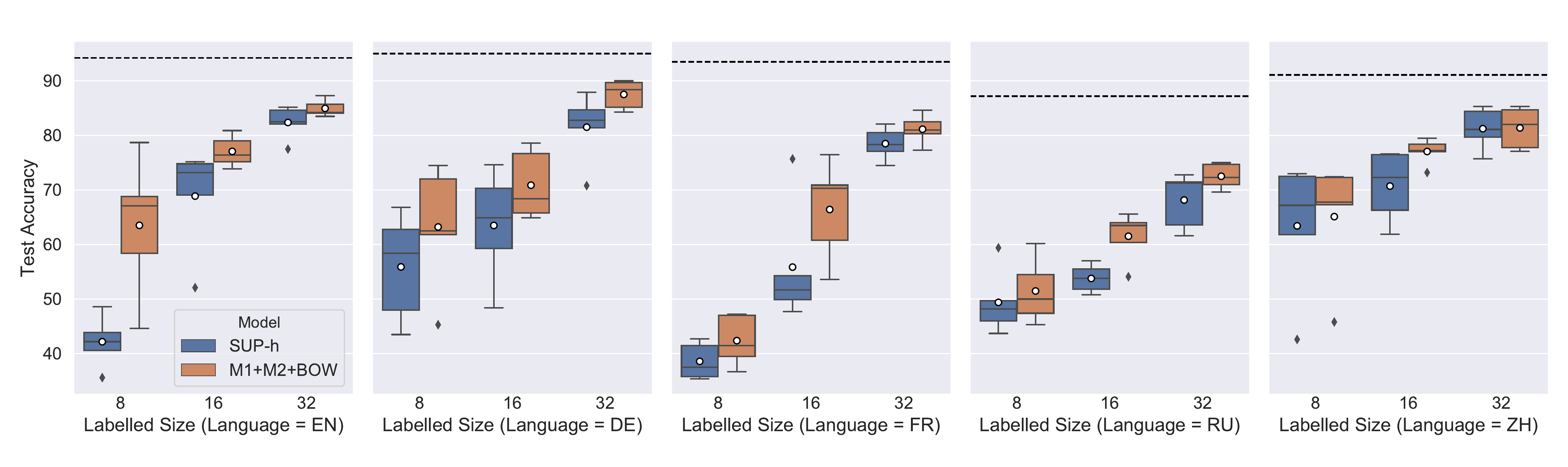}
	 \vspace{-2mm}
	\caption{Boxplot of test accuracy scores for SUP-h and M1+M2+BOW over 5 runs. The mean is shown as white dot. The dashed line is the test mean accuracy of SUP-h trained on 1k labelled data of the corresponding language.}\label{fig:bert_result}
    \vspace{-2mm}
\end{figure*}

\subsection{Multi-lingual BERT Encoder}\label{sec:bert}
\paragraph{Experimental Setup.}
We use the cased mBERT, a 12 layer Transformer \citep{NIPS2017_7181} trained on Wikipedia of 104 languages with 100k shared WordPiece vocabulary.
The training corpus is larger than Europarl by orders of magnitude, and high-resource languages account for most of the corpus.
We use the best SDGM setup (M1+M2+BOW \S\ref{sec:mono}), on top of mBERT encoder against the mBERT supervised model with a linear layer as classifier (SUP-h) in 5 representative languages (\textsc{en, de, fr, ru, zh}). We report the results over 5 runs due to the training instability of BERT \citep{DBLP:journals/corr/abs-2002-06305,DBLP:journals/corr/abs-2006-04884}.

\paragraph{Classification Results.}
Figure \ref{fig:bert_result} demonstrates that M1+M2+BOW outperforms the SOTA supervised mBERT (SUP-h) on average across all languages. This corroborates the effectiveness of our SDGM in leveraging unlabelled data within smaller labelled data regime, as well as its independence from encoding architecture.\footnote{Compared to smaller pretraining corpus (\S\ref{sec:mono}), we found that the representations pretrained on large corpus are less prune to overfit to the training instances of the task. We observe that training without the KL regularisation yields better performance for SDMGs+mBERT.}
As expected, the gap is generally larger with $8$ and $16$ labelled data, but reduces as the data size grows to $32$. 
The variance shows similar pattern, but with relatively large values because of the instability of mBERT.
Interestingly, the performance difference seems to be more notable in high-resource languages with more pretrained data, whereas in languages with fewer pretrained texts or vocabulary overlaps such as \textsc{ru} and \textsc{zh}, the two models achieve closer results.

%% file: 06-conclusion.tex
\section{Conclusion}
We bridged between multi-lingual pretraining and deep generative models to form a semi-supervised learning framework for document classification.
While outperforming SOTA supervised models in several settings, we showed that the benefits of SDGMs are orthogonal to the encoding architecture or pretraining procedure.
It opens up a new avenue for SDGMs in low-resource NLP by incorporating unlabelled data potentially from different domains and languages.
Our preliminary results in cross-lingual zero-shot setting with SDGMs+NXVAE are promising, and we will continue the exploration in this direction as future work.



%% file: appendix.tex
\section{Derivations of semi-supervised ELBOs}
We derive the full ELBOs of both labelled and unlabelled data for M1+M2 and Auxiliary Skip Deep Generative Model (AUX; \citet{DBLP:conf/icml/MaaloeSSW16}).\footnote{As mentioned in the footnote of original paper, we compare M1+M2 with AUX in LSTM encoder with VAE pretraining, but found that the simpler M1+M2 performs better. Results on AUX can be found in \S\ref{app:sec:mono}.}
We first use $(\boldsymbol{\cdot})$ to represent different conditional variables for the two models so that the derivations can be unified, then we will realise it with the model-specific conditions in the end.

As written in the paper, the labelled ELBO for both models is:
\begin{align*}
&\mathbb{E}_{q_{\phi}(\mathbf{z}_1, \mathbf{z}_2 | \mathbf{x}, y)}\Big[\!\log\frac{p_{\theta}(\mathbf{x}, y, \mathbf{z}_1, \mathbf{z}_2)}{q_{\phi}(\mathbf{z}_1, \mathbf{z}_2 | \mathbf{x}, y)}\Big]\!\\
&= \mathcal{L}(\mathbf{x}, y) \le \log p(\mathbf{x}, y)
\end{align*}
Expanding the ELBO, we will have:
{\footnotesize
\begin{align*}
&\mathbb{E}_{q_{\phi}(\mathbf{z}_1, \mathbf{z}_2 | \mathbf{x}, y)}[
\log \frac{p_{\theta}(\mathbf{x}, y, \mathbf{z}_1, \mathbf{z}_2)}{q_{\phi}(\mathbf{z}_1, \mathbf{z}_2 | \mathbf{x}, y)}
]\\
=&\mathbb{E}_{q_{\phi}(\mathbf{z}_1 | \mathbf{x})q_{\phi}(\mathbf{z}_2 | \boldsymbol{\cdot})}[\\
&\log p(\mathbf{z}_2) + \log p_{\theta}(\mathbf{z}_1 | \mathbf{z}_2, y) + \log p_{\theta}(\mathbf{x} | \boldsymbol{\cdot}) + \log p(y) -\\
&\log q_{\phi}(\mathbf{z}_2 | \boldsymbol{\cdot}) - \log q_{\phi}(\mathbf{z}_1 | \mathbf{x}) 
]\\
=&\mathbb{E}_{q_{\phi}(\mathbf{z}_1 | \mathbf{x})q_{\phi}(\mathbf{z}_2 | \boldsymbol{\cdot})}[\log p_{\theta}(\mathbf{x} | \boldsymbol{\cdot})] -\\
&\mathbb{E}_{q_{\phi}(\mathbf{z}_1 | \mathbf{x})q_{\phi}(\mathbf{z}_2 | \boldsymbol{\cdot})}[\\
&\log q_{\phi}(\mathbf{z}_2 | \boldsymbol{\cdot}) + \log q_{\phi}(\mathbf{z}_1 | \mathbf{x}) -\\
&\log p(\mathbf{z}_2) - \log p_{\theta}(\mathbf{z}_1 | \mathbf{z}_2, y) - \log p(y)
]\\
=&
\mathbb{E}_{q_{\phi}(\mathbf{z1} | \mathbf{x})
q_{\phi}(\mathbf{z}_2 | \boldsymbol{\cdot})}
[
\log p_{\theta}(\mathbf{x} | \boldsymbol{\cdot})
] -\\
&
\mathbb{E}_{q_{\phi}(\mathbf{z}_1 | \mathbf{x})q_{\phi}(\mathbf{z}_2 | \boldsymbol{\cdot})}[\\
&\log \frac{q_{\phi}(\mathbf{z}_2 | \boldsymbol{\cdot})}{p(\mathbf{z}_2)} + \log \frac{q_{\phi}(\mathbf{z}_1 | \mathbf{x})}{p_{\theta}(\mathbf{z}_1 | \mathbf{z}_2, y)} - \log p(y)
]
\end{align*}
}
After realising ($\boldsymbol{\cdot}$), we can then obtain the labelled ELBO for M1+M2 and AUX in the original paper:
{\footnotesize
\begin{align*}
&\mathcal{L}_{\textrm{M1+M2}}(\mathbf{x}, y)\\
=& \mathbb{E}_{q_{\phi}(\mathbf{z}_1 | \mathbf{x})}
[
\log p_{\theta}(\mathbf{x} | \mathbf{z}_1)
]
-\\
&\mathbb{E}_{q_{\phi}(\mathbf{z}_1 | \mathbf{x})q_{\phi}(\mathbf{z}_2 | \mathbf{z}_1, y)}[\\
&\log \frac{q_{\phi}(\mathbf{z}_2 | \mathbf{z}_1, y)}{p(\mathbf{z}_2)} 
+ 
\log \frac{q_{\phi}(\mathbf{z}_1 | \mathbf{x})}{p_{\theta}(\mathbf{z}_1 | \mathbf{z}_2, y)} 
- \log p(y)
]
\end{align*}
\begin{align*}
&\mathcal{L}_{\textrm{AUX}}(\mathbf{x}, y)\\
=&\mathbb{E}_{q_{\phi}(\mathbf{z}_1 | \mathbf{x})
q_{\phi}(\mathbf{z}_2 | \mathbf{z}_1, \mathbf{x}, y)}
[
\log p_{\theta}(\mathbf{x} | \mathbf{z}_1, \mathbf{z}_2, y)
]
-\\
&\mathbb{E}_{q_{\phi}(\mathbf{z}_1 | \mathbf{x})q_{\phi}(\mathbf{z}_2 | \mathbf{z}_1, \mathbf{x}, y)}[\\
&\log \frac{q_{\phi}(\mathbf{z}_2 | \mathbf{z}_1, \mathbf{x}, y)}{p(\mathbf{z}_2)} 
+ 
\log \frac{q_{\phi}(\mathbf{z}_1 | \mathbf{x})}{p_{\theta}(\mathbf{z}_1 | \mathbf{z}_2, y)} 
-
\log p(y)
]
\end{align*}
}

For the unlabelled ELBO, $y$ is unobservable:
\begin{align*}
&\mathbb{E}_{q_{\phi}(\mathbf{z}_1, \mathbf{z}_2, y | \mathbf{x})}\Big[\!\log\frac{p_{\theta}(\mathbf{x}, y, \mathbf{z}_1, \mathbf{z}_2)}{q_{\phi}(\mathbf{z}_1, \mathbf{z}_2, y | \mathbf{x})}\Big]\!\\ 
& = \mathcal{U}(\mathbf{x}) \le \log p(\mathbf{x})
\end{align*}
After expansion:
{\footnotesize
\begin{align*}
&\mathbb{E}_{q_{\phi}(\mathbf{z}_1, \mathbf{z}_2, y | \mathbf{x})}[
\log \frac{p_{\theta}(\mathbf{x}, y, \mathbf{z}_1, \mathbf{z}_2)}{q_{\phi}(\mathbf{z}_1, \mathbf{z}_2, y | \mathbf{x})}
]\\
=&\mathbb{E}_{q_{\phi}(\mathbf{z}_1 | \mathbf{x})q_{\phi}(y | \boldsymbol{\cdot})q_{\phi}(\mathbf{z}_2 | \boldsymbol{\cdot})}[\\
&\log p(\mathbf{z}_2) + \log p_{\theta}(\mathbf{z}_1 | \mathbf{z}_2, y) + \log p_{\theta}(\mathbf{x} | \boldsymbol{\cdot}) + \log p(y) -\\
&\log q_{\phi}(\mathbf{z}_2 | \boldsymbol{\cdot}) - \log q_{\phi}(\mathbf{z}_1 | \mathbf{x}) - \log q_{\phi}(y | \boldsymbol{\cdot})
]\\
=&\mathbb{E}_{q_{\phi}(\mathbf{z}_1 | \mathbf{x})q_{\phi}(y | \boldsymbol{\cdot})q_{\phi}(\mathbf{z}_2 | \boldsymbol{\cdot})}[\log p_{\theta}(\mathbf{x} | \boldsymbol{\cdot})] -\\
&\mathbb{E}_{q_{\phi}(\mathbf{z}_1 | \mathbf{x})q_{\phi}(y | \boldsymbol{\cdot})q_{\phi}(\mathbf{z}_2 | \boldsymbol{\cdot})}[\\
&\log q_{\phi}(\mathbf{z}_2 | \boldsymbol{\cdot}) + \log q_{\phi}(\mathbf{z}_1 | \mathbf{x}) + \log q_{\phi}(y | \boldsymbol{\cdot}) -\\
&\log p(\mathbf{z}_2) - \log p_{\theta}(\mathbf{z}_1 | \mathbf{z}_2, y) - \log p(y)
]\\
=&
\mathbb{E}_{q_{\phi}(\mathbf{z1} | \mathbf{x})
q_{\phi}(y | \boldsymbol{\cdot})q_{\phi}(\mathbf{z}_2 | \boldsymbol{\cdot})}
[
\log p_{\theta}(\mathbf{x} | \boldsymbol{\cdot})
] -\\
&
\mathbb{E}_{q_{\phi}(\mathbf{z}_1 | \mathbf{x})q_{\phi}(y | \boldsymbol{\cdot})q_{\phi}(\mathbf{z}_2 | \boldsymbol{\cdot})}[\\
&\log \frac{q_{\phi}(\mathbf{z}_2 | \boldsymbol{\cdot})}{p(\mathbf{z}_2)} + \log \frac{q_{\phi}(\mathbf{z}_1 | \mathbf{x})}{p_{\theta}(\mathbf{z}_1 | \mathbf{z}_2, y)} + \log \frac{q_{\phi}(y | \boldsymbol{\cdot})}{p(y)}
]
\end{align*}
}
Similarly, we will get unlabeled ELBO of M1+M2 and AUX:
{\footnotesize
\begin{align*}
&\mathcal{U}_{\textrm{M1+M2}}(\mathbf{x})\\
=&\mathbb{E}_{q_{\phi}(\mathbf{z}_1 | \mathbf{x})}[\log p_{\theta}(\mathbf{x} | \mathbf{z}_1)] -\\
&\mathbb{E}_{q_{\phi}(\mathbf{z}_1 | \mathbf{x})q_{\phi}(y | \mathbf{z}_1)q_{\phi}(\mathbf{z}_2 | \mathbf{z}_1, y)}[\\
&\log \frac{q_{\phi}(\mathbf{z}_2 | \mathbf{z}_1, y)}{p(\mathbf{z}_2)} + 
\log \frac{q_{\phi}(\mathbf{z}_1 | \mathbf{x})}{p_{\theta}(\mathbf{z}_1 | \mathbf{z}_2, y)} +
\log \frac{q_{\phi}(y | \mathbf{z}_1)}{p(y)}
]
\end{align*}
\begin{align*}
&\mathcal{U}_{\textrm{AUX}}(\mathbf{x})\\
=&\mathbb{E}_{q_{\phi}(\mathbf{z}_1 | \mathbf{x})q_{\phi}(y | \mathbf{z}_1, \mathbf{x})q_{\phi}(\mathbf{z}_2 | \mathbf{z}_1, \mathbf{x}, y)}[\log p_{\theta}(\mathbf{x} | \mathbf{z}_1, \mathbf{z}_2, y)] -\\
&\mathbb{E}_{q_{\phi}(\mathbf{z}_1 | \mathbf{x})q_{\phi}(y | \mathbf{z}_1, \mathbf{x})q_{\phi}(\mathbf{z}_2 | \mathbf{z}_1, \mathbf{x}, y)}[\\
&\log \frac{q_{\phi}(\mathbf{z}_2 | \mathbf{z}_1, \mathbf{x}, y)}{p(\mathbf{z}_2)} + 
\log \frac{q_{\phi}(\mathbf{z}_1 | \mathbf{x})}{p_{\theta}(\mathbf{z}_1 | \mathbf{z}_2, y)} + 
\log \frac{q_{\phi}(y | \mathbf{z}_1, \mathbf{x})}{p(y)}
]
\end{align*}
}

In our experiments, we sample $\mathbf{z}_1$ and $\mathbf{z}_2$ once during inference, so both labeled and unlabeled ELBOs can be approximated by:
{\footnotesize
\begin{align*}
&\mathcal{L}(\mathbf{x}, y)\\
=&
\mathbb{E}_{q_{\phi}(\mathbf{z}_1 | \mathbf{x})
q_{\phi}(\mathbf{z}_2 | \boldsymbol{\cdot})
}
[
\log p_{\theta}(\mathbf{x} | \boldsymbol{\cdot})
] -\\
&
\mathbb{E}_{q_{\phi}(\mathbf{z}_1 | \mathbf{x})q_{\phi}(\mathbf{z}_2 | \boldsymbol{\cdot})}[\\
&\log \frac{q_{\phi}(\mathbf{z}_2 | \boldsymbol{\cdot})}{p(\mathbf{z}_2)} + \log \frac{q_{\phi}(\mathbf{z}_1 | \mathbf{x})}{p_{\theta}(\mathbf{z}_1 | \mathbf{z}_2, y)} - \log p(y)
]\\
\approx & \log p_{\theta}(\mathbf{x} | \boldsymbol{\cdot}) + \log p(y) -\\
& \textrm{KL}(q_{\phi}(\mathbf{z}_2 | \boldsymbol{\cdot}) \| p(\mathbf{z}_2)) - 
\textrm{KL}(q_{\phi}(\mathbf{z}_1 | \mathbf{x}) \| p_{\theta}(\mathbf{z}_1 | \mathbf{z}_2, y))
\end{align*}
\begin{align*}
&\mathcal{U}(\mathbf{x})\\
=&
\mathbb{E}_{q_{\phi}(\mathbf{z}_1 | \mathbf{x})
q_{\phi}(y | \boldsymbol{\cdot})q_{\phi}(\mathbf{z}_2 | \boldsymbol{\cdot})
}
[
\log p_{\theta}(\mathbf{x} | \boldsymbol{\cdot})
] -\\
&
\mathbb{E}_{q_{\phi}(\mathbf{z}_1 | \mathbf{x})q_{\phi}(y | \boldsymbol{\cdot})q_{\phi}(\mathbf{z}_2 | \boldsymbol{\cdot})}[\\
&\log \frac{q_{\phi}(\mathbf{z}_2 | \boldsymbol{\cdot})}{p(\mathbf{z}_2)} + \log \frac{q_{\phi}(\mathbf{z}_1 | \mathbf{x})}{p_{\theta}(\mathbf{z}_1 | \mathbf{z}_2, y)} + \log \frac{q_{\phi}(y | \boldsymbol{\cdot})}{p(y)}
]\\
\approx & \log p_{\theta}(\mathbf{x} | \boldsymbol{\cdot}) - \textrm{KL}(q_{\phi}(y | \boldsymbol{\cdot}) \| p(y)) - \\
&\mathbb{E}_{q_{\phi}(y | \boldsymbol{\cdot})}[\textrm{KL}(q_{\phi}(\mathbf{z}_2 | \boldsymbol{\cdot}) \| p(\mathbf{z}_2))] -\\
&\mathbb{E}_{q_{\phi}(y | \boldsymbol{\cdot})}[\textrm{KL}(q_{\phi}(\mathbf{z}_1 | \mathbf{x}) \| p_{\theta}(\mathbf{z}_1 | \mathbf{z}_2, y))]
\end{align*}
}

\section{Factorisation of M1+M2 and AUX}
The two models have different factorisations, with M1+M2 being written as:
\begin{equation*}
\setlength{\abovedisplayskip}{3pt}
\setlength{\belowdisplayskip}{3pt}
\setlength{\abovedisplayshortskip}{3pt}
\setlength{\belowdisplayshortskip}{3pt}
\resizebox{0.95\linewidth}{!}{
  \begin{minipage}{\linewidth}
\centering
$
\begin{aligned}
q_{\phi}(\mathbf{z}_1, \mathbf{z}_2 | \mathbf{x}, y) &= q_{\phi}(\mathbf{z}_1 | \mathbf{x})q_{\phi}(\mathbf{z}_2 | \mathbf{z}_1, y)\\
q_{\phi}(\mathbf{z}_1, \mathbf{z}_2, y | \mathbf{x}) &= q_{\phi}(\mathbf{z}_1 | \mathbf{x})q_{\phi}(y | \mathbf{z}_1)q_{\phi}(\mathbf{z}_2 | \mathbf{z}_1, y)\\
p_{\theta}(\mathbf{x}, y, \mathbf{z}_1, \mathbf{z}_2) &= p(y)p(\mathbf{z}_2)p_{\theta}(\mathbf{z}_1 | \mathbf{z}_2, y)p_{\theta}(\mathbf{x} | \mathbf{z}_1)\\
\mathcal{J}_{cls}(\mathbf{x}, y) &= \mathbb{E}_{q_{\phi}(\mathbf{z}_1|\mathbf{x})}[q_{\phi}(y|\mathbf{z}_1)]
\end{aligned}
$
\end{minipage}
}
\end{equation*}
and AUX is factorised as follows:
\begin{equation*}
\setlength{\abovedisplayskip}{3pt}
\setlength{\belowdisplayskip}{3pt}
\setlength{\abovedisplayshortskip}{3pt}
\setlength{\belowdisplayshortskip}{3pt}
\resizebox{0.85\linewidth}{!}{
  \begin{minipage}{\linewidth}
\centering
$
\begin{aligned}
q_{\phi}(\mathbf{z}_1, \mathbf{z}_2 | \mathbf{x}, y) &= q_{\phi}(\mathbf{z}_1 | \mathbf{x})q_{\phi}(\mathbf{z}_2 | \mathbf{z}_1, \mathbf{x}, y)\\
q_{\phi}(\mathbf{z}_1, \mathbf{z}_2, y | \mathbf{x}) &= q_{\phi}(\mathbf{z}_1 | \mathbf{x})q_{\phi}(y | \mathbf{z}_1, \mathbf{x})q_{\phi}(\mathbf{z}_2 | \mathbf{z}_1, \mathbf{x}, y)\\
p_{\theta}(\mathbf{x}, y, \mathbf{z}_1, \mathbf{z}_2) &= p(y)p(\mathbf{z}_2)p_{\theta}(\mathbf{z}_1 | \mathbf{z}_2, y)p_{\theta}(\mathbf{x} | \mathbf{z}_1, \mathbf{z}_2, y)\\
\mathcal{J}_{cls}(\mathbf{x}, y) &= \mathbb{E}_{q_{\phi}(\mathbf{z}_1|\mathbf{x})}[q_{\phi}(y|\mathbf{z}_1, \mathbf{x})]
\end{aligned}
$
\end{minipage}
}
\end{equation*}
where $q_{\phi}(\mathbf{z}_1 | \mathbf{x})$, $q_{\phi}(\mathbf{z}_2 | \boldsymbol{\cdot})$, and $p_{\theta}(\mathbf{z}_1 | \mathbf{z}_2, y)$ are parameterised as diagonal Gaussians, and other distributions are defined as: 
\begin{equation*}
\setlength{\abovedisplayskip}{3pt}
\setlength{\belowdisplayskip}{3pt}
\setlength{\abovedisplayshortskip}{3pt}
\setlength{\belowdisplayshortskip}{3pt}
\resizebox{0.95\linewidth}{!}{
  \begin{minipage}{\linewidth}
\centering
$
\begin{aligned}
q_{\phi}(y | \boldsymbol{\cdot}) &= \textrm{Cat}(y | \pi_{\phi}(\boldsymbol{\cdot}))\hspace{11mm}
p(y) = \textrm{Cat}(y|\pi)\\
p(\mathbf{z}_2) &= \mathcal{N}(\mathbf{z}_2 | \mathbf{0}, \mathbf{I})\hspace{10mm}
p_{\theta}(\mathbf{x} | \boldsymbol{\cdot}) = f(\mathbf{x}, \boldsymbol{\cdot}; \theta)
\end{aligned}
$
\end{minipage}}
\end{equation*}
where $\textrm{Cat}(\cdot)$ is a multinomial distribution and $y$ is treated as latent variables if it is unobserved in unlabelled case.
$f(\mathbf{x}, \boldsymbol{\cdot}; \theta)$ serves as the decoder and calculates the likelihood of the input sequence $\mathbf{x}$. 

\section{Details on LSTM Encoder with VAE Pretraining}\label{app:pretraining}
\subsection{Data preprocessing and statistics}
We use two pairs of data from Europarl v7 \cite{Koehn2002EuroparlAM}:\footnote{\url{https://www.statmt.org/europarl/}.}
\textsc{en}-\textsc{de} and \textsc{en}-\textsc{fr}, which consist of four datasets in total: \textsc{en}$_{\textrm{\textsc{en}-\textsc{de}}}$, \textsc{de}$_{\textrm{\textsc{en}-\textsc{de}}}$, \textsc{en}$_{\textrm{\textsc{en}-\textsc{fr}}}$, and \textsc{fr}$_{\textrm{\textsc{en}-\textsc{fr}}}$.
Regarding \textsc{de}-\textsc{fr} data, we take the datasets of \textsc{de}$_{\textrm{\textsc{en}-\textsc{de}}}$ and \textsc{fr}$_{\textrm{\textsc{en}-\textsc{fr}}}$. 

For each language pair, the sentences in the same line of both datasets are a pair of parallel sentences.
We do the following preprocessing to each dataset: tokenization; lower case; substitute digits with 0; remove all punctuations; remove redundant spaces and empty lines.
Then we trim all four datasets into exactly the same sentence size.
We randomly split a small part of parallel sentences to build a dev. set, which leads to 189m lines of training set and 13995 lines of dev. set for each language.
Then we shuffle each dataset so that each language pair is not parallel anymore (for both train and dev. sets).

Our goal is to merge the two datasets of each pair and scramble them to form a single dataset.
In practice, we keep each dataset separate, and sample a batch randomly from one language alternatively during pretraining, so that the data from both languages are mixed.

\subsection{Model and training details}
Instead of optimising the standard VAE, we optimise the following objective for NXVAE \cite{DBLP:conf/iclr/HigginsMPBGBML17,DBLP:conf/icml/AlemiPFDS018}:
\begin{equation}\label{app:eq:vae}
\setlength{\abovedisplayskip}{3pt}
\setlength{\belowdisplayskip}{3pt}
\setlength{\abovedisplayshortskip}{3pt}
\setlength{\belowdisplayshortskip}{3pt}
\resizebox{0.8\linewidth}{!}{
  \begin{minipage}{\linewidth}
$
\begin{aligned}
\mathcal{J}(\mathbf{x}) &= \mathbb{E}_{q_{\phi}(\mathbf{z} | \mathbf{x})}[\log p_{\theta}(\mathbf{x} | \mathbf{z})] - \alpha \textrm{KL}(q_{\phi}(\mathbf{z} | \mathbf{x}) \| p(\mathbf{z}))
\end{aligned}
$
\end{minipage}}
\end{equation}
where we manually tune the fixed hyperparameter $\alpha$ on \textsc{en}-\textsc{de} data to reach a good balance between the reconstruction and the KL empirically.
We select $\alpha = 0.1$ and apply it for the pretraining of other language pairs as well.
The model and training details of XNVAE are shown in Table \ref{app:tb:nxvae_hyp}~(left).

\subsection{Pretraining other models}\label{sec:app_other_models}
For MLDoc supervised document classification, we also pretrain other baseline models to compare with ONLY for \textbf{\textsc{en-de}} pair: 

\paragraph{Cross-lingual VAE with parallel input (PEMB; \citet{ijcai2017-582}):}
For the model of \citet{ijcai2017-582}, we run the original code directly on the same \textsc{en-de} Europarl data without changing any of the model architecture.
Since the model requires parallel input, we take the preprocessed and split \textsc{en}-\textsc{de} data.
However, we do not shuffle each dataset, but rather feed them as parallel input to the model, so that the model and our corresponding NXVAE use the same amount and content of the data. 

\paragraph{Subword-based non-parallel cross-lingual VAE SNXVAE:}
Instead of having separate vocabulary and decoders for each language, we use a single vocabulary and decoder for SNXVAE.
We build the vocabulary with SentencePiece\footnote{\url{https://github.com/google/sentencepiece}.} of size $1\textrm{e}4$.
All other settings are the same as NXVAE.
Its model and training details can be found in Table \ref{app:tb:nxvae_hyp}~(right).

\paragraph{Word and subword-based BERT model BERTW/BERTSW}:
For BERTW, we change the vocabulary and model size to be comparable with NXVAE. Note that the vocabulary size of BERTW is the same as the intersected vocabulary size of the two languages in NXVAE.
We only use the masked language model objective during pretraining, and discard the objective of next sentence prediction.\footnote{Both word and subword-based models are trained with: \url{https://github.com/google-research/bert}.}
For BERTSW, we use the same vocabulary as SNXVAE and set the model to similar parameter size as SNXVAE.
The model and training details of BERTW and BERTSW are shown in Table \ref{app:tb:bert_hyp}.



\section{More Results on Document Classification}\label{app:sec:mono}

\subsection{LSTM Encoder with VAE Pretraining}
\paragraph{Supervised Learning.} Our base model is NXVAE-z$_1$, which adds an MLP classifier $q_{\phi}(y | \mathbf{z}_1)$ on top of the encoder with the same architecture of the NXVAE.
The similar applies to the subword-based models SNXVAE-z$_1$.
NXVAE-h takes the deterministic $\mathbf{h}$ as the input to $q_{\phi}(y | \mathbf{x})$.
All our baseline models with pretrained embeddings use the architecture of NXVAE-z$_1$.
For fastText (FT), we train the embeddings of both languages with the same data of 
\textsc{en}$_{\textrm{\textsc{en}-\textsc{de}}}$ and \textsc{de}$_{\textrm{\textsc{en}-\textsc{de}}}$.
For MUSE, we align on the pretrained FT embeddings.
For BERTW and BERTSW, we use the library Transformers\footnote{\url{https://github.com/huggingface/transformers}.} for classification, and initialise the models with the corresponding pretrained parameters.
All model and training details can be found in Table \ref{app:tb:mldoc_sup_hyp}.
The comparison results of word-based and subword-based models are shown in Table \ref{app:tb:wordvssw}.

\paragraph{Semi-supervised learning with SDGMs.}
The main model (NXVAE) and training details are the same as in supervised learning.
Besides M1+M2, we also compare with AUX \citep{DBLP:conf/icml/MaaloeSSW16} with the two decoder types.
The training details are shown in Table \ref{app:tb:semi_hyp}.
Regarding the decoding of GRU, all conditional latent variables of $p_{\theta}(\mathbf{x} | \boldsymbol{\cdot})$ are fed as extra input at each decoding step \cite{DBLP:conf/aaai/XuSDT17}. 

We tune all semi-supervised models on \textsc{en}$_{\textrm{\textsc{en-de}}}$ with 32 labels in semi-supervised settings, and then apply it to all other languages and data sizes.
We tune only one hyperparameter: the scaling factor $\beta$ in the weight for the classification loss $\alpha$ in the original SDGM paper \cite{DBLP:conf/icml/MaaloeSSW16}:
\begin{align*}
\alpha = \beta \frac{N_l + N_u}{N_l}
\end{align*}
where $N_l$ and $N_u$ are labelled and unlabelled data point numbers.
We tune $\beta$ from $\{0.1, 0.2, 0.5, 1.0, 2.0, 5.0, 10.0, 20.0\}$.
We pick the $\beta$ with the best dev. performance for each model, and randomly select one when there is a tie. 
Then we use such fixed $\beta$ for all other experiments across different training data sizes and languages.

The results of AUX can be seen in Table \ref{app:tb:auxmono} along with M1+M2 results from the original paper.
The parameter size of each model is shown in Table \ref{app:tb:monops}.

\subsection{mBERT Encoder}
The supervised model (SUP-h) adds a single linear transformation layer on the pooled \texttt{[CLS]} representation of mBERT, and M1+M2+BOW adds the corresponding SDGM framework on the same mBERT output.
Like BERT, as mBERT uses a shared WordPiece vocabulary across languages, the parameter size of the same model will be the same for each language.
All model and training details along with parameter size can be found in Table \ref{app:tb:mbertsetting}.

For tuning the hyperparameter of M1+M2+BOW, different from LSTM encoder with VAE pretraining, we set $\alpha$ fixed to $\alpha = \beta$.
We tune $\beta$ on \textsc{en} with 8 labels in semi-supervised settings with 5 trials from $\{0.1, 0.2, 0.5, 1.0, 2.0, 5.0, 10.0, 20.0, 50.0\}$, pick the $\beta$ with the best \textbf{average} dev. performance, and then apply it to all other languages and data sizes.
We report the mean and variance over 5 trials, and the full results for both models can be seen in Table \ref{app:tb:mbertmono}.

\section{Conditional document generation}
Semi-supervised deep generative models can not only explore the complex data distributions, but are also equipped with the ability to generate documents conditioned on latent codes, which is another advantage over other semi-supervised models.
We follow \citet{NIPS2014_5352} by varying latent variable $y$ for generation, and fixing $\mathbf{z}_2$ either sampled from the prior (Table \ref{app:tb:ucgen}) or obtained from the input through the inference model (Table \ref{app:tb:cgen}), and generate sequence samples from the trained semi-supervised models M1+M2+BOW and M1+M2+GRU.\footnote{All models are treined on \textsc{en}$_{\textrm{en-fr}}$ with 128 labelled data.}

Overall, all models generate words or utterances directly related to the class, with the class labels among top nouns generated by BOW models, and subjects/objects in sentences from GRU are also pertaining to corresponding classes.  
However, we also observe that the utterances in GRU are not fluent with many repetitions.
We argue that it is caused by the high proportion of UNK in the training corpus that makes the sequence generation harder, supported by the fact that the most probable word in all BOW decoders is always UNK.

\begin{table*}[t]
	\centering
    \def\arraystretch{0.95}
    {\footnotesize
	\begin{tabularx}{\textwidth}{l cc}
	    Hyperparameter & NXVAE & SNXVAE\\\toprule
	    vocabulary size & 4e4 (\textsc{en}), 5e4 (\textsc{de}, \textsc{fr}) & 1e4\\
	    embedding size  & 300 & 300\\
	    embedding dropout & 0.2 & 0.2\\
	    encoder & BiLSTM & BiLSTM\\
	    encoder input dimension & 300 & 300\\
	    encoder hidden dimension & 600 for each direction & 600 for each direction\\
	    encoder layer number & 2 & 2\\
	    encoder dropout & 0.2 & 0.2\\
	    discriminator configuration & [2 $\times$ 600, 1024, leakyrelu, 1024, 1] & [2 $\times$ 600, 1024, leakyrelu, 1024, 1]\\
	    inferer ($\mathbf{h}$ to $\mathbf{\mu}$ or $\mathbf{\log \sigma}$) configuration & [2 $\times$ 600, 300, batchNorm, relu, 300] & [2 $\times$ 600, 300, batchNorm, relu, 300]\\
	    $\mathbf{z}$ dimension & 300 & 300\\
	    parameter size & 41.8M (\textsc{en-de} and \textsc{en-fr})/ 44.9M (\textsc{de-fr}) & 17.8M\\
	    running time & $\sim$ 1 day & less than 1 day\\
	    tie embeddings of encoder and decoder & True & True\\
	    sentence length threshold &  median length of training data & median length of training data\\
	    $\alpha$ in Equation \ref{app:eq:vae} & $\{$0.01, 0.02, 0.05, \textbf{0.1}, 0.2, 0.5, 1.0$\}$ & 0.1\\
	    training epoch & 500 & 500\\
	    early stopping & 5 epochs on dev. negative likelihood & 5 epochs\\
	    batch size & 128 & 128 \\
	    validate on dev. & every 4000 iterations & every 4000 iterations\\
	    optimiser & Adam & Adam\\ 
	    learning rate & 5e-4 & 5e-4\\
		\bottomrule       
	\end{tabularx}}%
    \vspace{-2mm}
     \caption{Model and training details of NXVAE.} \label{app:tb:nxvae_hyp}
\end{table*}

\begin{table*}[t]
	\centering
    \def\arraystretch{0.95}
    {\footnotesize
	\begin{tabularx}{\columnwidth}{l cc}
	Hyperparameter & \textsc{bertw} & \textsc{bertsw}\\\toprule
    vocabulary size & 84101 & 10005\\
    hidden size & 300 & 300\\
    max position embeddings & 512 & 512\\
    hidden dropout prob & 0.1 & 0.1\\
    hidden activation & gelu & gelu\\
    intermediate size & 2100 & 1800\\
    num attention heads & 12 & 12\\
    attention probs dropout prob & 0.1 & 0.1\\
    num hidden layers & 12 & 11\\
    parameter size &  45.0M & 19.1M\\
    running time & $\sim$ 5 days & $\sim$ 3 days\\\midrule
    max seq length & \multicolumn{2}{c}{200}\\
    max predictions per seq & \multicolumn{2}{c}{30}\\
    masked lm prob & \multicolumn{2}{c}{0.15}\\
    batch size & \multicolumn{2}{c}{32}\\
    optimiser & \multicolumn{2}{c}{Adam}\\
    learning rate & \multicolumn{2}{c}{1e-4}\\
    weight decay & \multicolumn{2}{c}{0.01}\\
    num train steps & \multicolumn{2}{c}{1e6}\\
	\bottomrule       
	\end{tabularx}}%
    \vspace{-2mm}
     \caption{Model and training details of BERTW and BERTSW.} \label{app:tb:bert_hyp}
\end{table*}

\begin{table*}[t]
	\centering
    \def\arraystretch{0.95}
    {\footnotesize
	\begin{tabularx}{\textwidth}{l ll}
	Hyperparameter & BERTW/BERTSW & \textsc{VAE}-based\\\toprule
    vocabulary     & same as pretrained model & same as pretrained model\\
    training epoch & 5000 & 5000\\
    early stopping & 1000 epochs on dev. accuracy & 1000\\
    batch size     & 16 & 16\\
    running time   & $\sim$5.5h & $\sim$2.5h\\
    sentence length & 200 & 200\\
    optimiser & Adam & Adam\\
    learning rate & 2e-5 & 5e-4\\
    classifier & [input\_dim, class\_num] & [input\_dim, 1024, leakyrelu, 1024, class\_num]\\
	\bottomrule       
	\end{tabularx}}%
    \vspace{-2mm}
     \caption{LSTM encoder with VAE pretraining: model and training details of MLDoc supervised document classification. The running time is calculated on \textsc{en}$_{\textsc{en-de}}$ with 32 labelled data for all models.} \label{app:tb:mldoc_sup_hyp}
\end{table*}

\begin{table*}[t]
	\centering
    \footnotesize
    \setlength{\tabcolsep}{2.0pt}
	\begin{tabularx}{\columnwidth}{@{}l cccc cccc}
	\toprule
	
    \textbf{\textsc{en-de}} & \multicolumn{4}{c}{\textsc{en}} & \multicolumn{4}{c}{\textsc{de}}\\ 
    \cmidrule[1.5pt](lr){2-5}\cmidrule[1.5pt](lr){6-9}
    
	& \textsc{32} & \textsc{64} & \textsc{128} & \textsc{full} &
	\textsc{32} & \textsc{64} & \textsc{128} & \textsc{full}\\ \toprule 
	
    \textsc{bertw}       & 67.7            & 72.7          & \textbf{84.6} & \textbf{91.8} & 58.1          & 77.5          & \textbf{89.2}             & 94.0\\ 
    \textsc{bertsw}      & 54.4            & 69.0          & 83.0          & 91.4          & 62.2          & 80.1          & 84.3                      & \textbf{94.1} \\
    \cmidrule(lr){2-5}\cmidrule(lr){6-9}
    
    \textsc{nxvae-}z$_1$      & 63.9            & 71.4          & 77.0          & 91.6          & 65.0          & 73.8          & 82.7                      & 93.0\\
    \textsc{snxvae-}z$_1$     & 68.9            & \textbf{76.5} & 79.2          & 90.3          & 69.0          & 79.4          & 85.5                      & 91.7\\

	\bottomrule
	\end{tabularx}
    \vspace{-2mm}
     \caption{LSTM encoder with VAE pretraining: comparisons of word-based models and subword-based models for BERT and NXVAE in MLDoc supervised document classification. Word-based results are directly from the original paper.} \label{app:tb:wordvssw}
     \vspace{-3mm}
\end{table*}

\begin{table*}[t]
	\centering
    \def\arraystretch{0.95}
    {\footnotesize
	\begin{tabularx}{\textwidth}{l lllll}
	Hyperparameter & M1+M2 & M1+M2+BOW  & M1+M2+GRU & AUX+BOW & AUX+GRU\\\toprule
    training epoch & 5000 & 5000 & 5000 & 5000 & 5000\\
    early stopping & 1000 & 1000 & 1000 & 1000 & 1000\\
    best $\beta$   & 0.1 & 0.2 & 10.0 & 20.0 & 5.0\\
    $\mathbf{z}_1$ dim & 300 & 300 & 300 & 300 & 300\\
    $\mathbf{z}_2$ dim & 300 & 300 & 300 & 300 & 300\\
    tie embedding & - & False & False & False & False\\
    running time & $\sim$2h & $\sim$12h & $\sim$14h & $\sim$13h & $\sim$ 14.5h\\
    GRU input dim & - & - & 100 & - & 100\\
    GRU hidden dim & - & - & 50 & - & 50\\
    GRU layers & - & - & 1 & - & 1\\
    GRU dropout prob & - & - & 0.5 & - & 0.5\\
	\bottomrule       
	\end{tabularx}}%
    \vspace{-2mm}
     \caption{LSTM encoder with VAE pretraining: model and training details of MLDoc semi-supervised document classification. The running time is calculated on \textsc{en}$_{\textsc{en-de}}$ with 32 labelled data for all models.} \label{app:tb:semi_hyp}
\end{table*}

\begin{table*}[t]
	\centering
    \footnotesize
    \setlength{\tabcolsep}{2pt}
	\begin{tabularx}{\columnwidth}{@{}l cccc cccc}
	\toprule
	   \textbf{\textsc{en-de}} & \multicolumn{4}{c}{\textsc{en}} & \multicolumn{4}{c}{\textsc{de}}\\ 
	    \cmidrule[1.5pt](lr){2-5}\cmidrule[1.5pt](lr){6-9}
	    		& \textsc{32} & \textsc{64} & \textsc{128} & \textsc{full} &
		\textsc{32} & \textsc{64} & \textsc{128} & \textsc{full}\\ \toprule 
			    \textsc{m1+m2}      & 56.6          & 67.1          & 70.3          & -    & 52.6           & 67.2          & 76.8          & -\\
	    \textsc{m1+m2+bow}  & \textbf{79.8} & \textbf{81.7} & 87.2          & -    & 70.5           & 79.6          & \textbf{89.7} & -\\ 
	    \textsc{m1+m2+gru}  & 75.3          & 79.4          & 84.9          & -    & 75.1           & 80.0          & 87.1          & -\\
	    \textsc{aux+bow}    & 78.8          & \textbf{81.7} & \textbf{87.7} & -    & \textbf{75.2}  & \textbf{86.2} & 89.3          & - \\
	    \textsc{aux+gru}    & 74.8          & 80.0          & 85.1          & -    & 72.2           & 76.5          & 87.6          & - \\
	\bottomrule
	
	   \textbf{\textsc{en-fr}} & \multicolumn{4}{c}{\textsc{en}} & \multicolumn{4}{c}{\textsc{fr}}\\ 
	    \cmidrule[1.5pt](lr){2-5}\cmidrule[1.5pt](lr){6-9}
	    		& \textsc{32} & \textsc{64} & \textsc{128} & \textsc{full} &
		\textsc{32} & \textsc{64} & \textsc{128} & \textsc{full}\\ \toprule 
		  \textsc{m1+m2}      & 71.8          & 73.5          & 76.5          & -    & 66.2           & 78.7          & 79.7          & -\\
	    \textsc{m1+m2+bow}  & \textbf{81.0} & \textbf{85.5} & \textbf{88.2} & -    & 80.3           & \textbf{86.0} & \textbf{88.8} & -\\ 
	    \textsc{m1+m2+gru}  & 75.3          & 81.4          & 83.8          & -    & \textbf{80.7}  & 82.3          & 87.4          & -\\ 
	    \textsc{aux+bow}    & 79.8          & 83.4          & 87.1          & -    & 80.4           & 85.7          & 88.1          & -\\
	    \textsc{aux+gru}    & 78.3          & 81.3          & 86.6          & -    & \textbf{80.7}  & 83.2          & 85.4          & -\\ 
	\bottomrule

	   \textbf{\textsc{de-fr}} & \multicolumn{4}{c}{\textsc{de}} & \multicolumn{4}{c}{\textsc{fr}}\\ 
	    \cmidrule[1.5pt](lr){2-5}\cmidrule[1.5pt](lr){6-9}
	    		& \textsc{32} & \textsc{64} & \textsc{128} & \textsc{full} &
		\textsc{32} & \textsc{64} & \textsc{128} & \textsc{full}\\ \toprule 
	    \textsc{m1+m2}      & 59.1          & 70.6          & 75.4          & -    & 48.5           & 57.4          & 60.7          & -\\ 
	    \textsc{m1+m2+bow}  & \textbf{78.0} & \textbf{83.2} & 88.3          & -    & \textbf{81.4}  & \textbf{84.5} & \textbf{88.4} & -\\ 
	    \textsc{m1+m2+gru}  & 74.6          & 80.5          & 86.2          & -    & 66.2           & 77.2          & 81.9          & -\\  
	    \textsc{aux+bow}    & 74.6          & 82.9          & \textbf{89.0} & -    & 73.9           & 79.5          & 82.1          & -\\
	    \textsc{aux+gru}    & 70.7          & 79.5          & 80.3          & -    & 67.3           & 81.0          & 83.6          & -\\ 
	\bottomrule
	\end{tabularx}
    \vspace{-2mm}
     \caption{LSTM encoder with VAE pretraining: test accuracy of AUX models. The header numbers denote number of labelled training data instances. The best results are in bold. Other results related to M1+M2 are directly from the original paper.} \label{app:tb:auxmono}
\end{table*}

\begin{table*}[t]
	\centering
    \def\arraystretch{0.95}
    {\footnotesize
	\begin{tabularx}{\columnwidth}{l ccc}
	                        & \textsc{en} & \textsc{de} & \textsc{fr}\\\toprule
	    \textsc{embedding models} & 25.8M & 28.8M & 28.8M\\ 
	    \textsc{nxvae-}h          & 26.8M & 29.8M & 29.8M\\
	    \textsc{nxvae-}z$_1$     & 25.8M & 28.8M & 28.8M\\
	    \textsc{snxvze-}z$_1$    & 16.8M & 16.8M & 16.8M\\
	    \textsc{bertw}       & 45.0M & 45.0M & 45.0M\\
	    \textsc{bertsw}      & 19.1M & 19.1M & 19.1M\\\midrule
	    \textsc{m1+m2}       & 0.9M & 0.9M & 0.9M\\
	    \textsc{m1+m2+bow}   & 38.5M & 44.5M & 44.5M\\
	    \textsc{m1+m2+gru}   & 43.2M & 49.2M & 49.2M\\
	    \textsc{aux+bow}     & 43.8M & 49.8M & 49.8M\\
	    \textsc{aux+gru}     & 48.5M & 54.5M & 54.5M\\
		\bottomrule       
	\end{tabularx}}
    \vspace{-2mm}
     \caption{LSTM encoder with VAE pretraining: parameter size of all supervised and semi-supervised models. The difference between NXVAE-based models and BERTW is caused by language specific vocabulary of NXVAE, where only one vocabulary is used for \textbf{mono-lingual} document classification.} \label{app:tb:monops}
\end{table*}

\begin{table*}[t]
	\centering
    \def\arraystretch{0.95}
    {\footnotesize
	\begin{tabularx}{\textwidth}{l ll}
	Hyperparameter & SUP-h & M1+M2+BOW\\\toprule
    vocabulary size   & 1e5 & 1e5\\
    $\mathbf{z}_1$ dim & 768 & 768\\
    $\mathbf{z}_2$ dim & 768 & 768\\
    tie embedding & True & True\\
    best $\beta$   & - & 10.0\\
    training epoch & 500 & 500\\
    early stopping & 100 epochs on dev. accuracy & 100\\
    batch size     & 4 & 4\\
    running time   & $\sim$1h & $\sim$5h\\
    sentence length & 200 & 200\\
    optimiser & Adam & Adam\\
    learning rate & 2e-5 & 2e-5\\
    classifier & [768, class\_num] & [768, class\_num]\\
    parameter size &  178M & 185M\\
	\bottomrule       
	\end{tabularx}}%
    \vspace{-2mm}
     \caption{mBERT encoder: model and training details of MLDoc document classification. The running time is calculated on \textsc{en}$_{\textsc{en-de}}$ with 8 labelled data for both models.} \label{app:tb:mbertsetting}
\end{table*}

	                                            
	    
	                                          

\begin{table*}[t]
	\centering
    \footnotesize
	\begin{tabularx}{\linewidth}{ll XXXX} 
	\toprule
	& Model & 8 & 16 & 32 & \textsc{1k}\\\toprule
	
	\parbox[t]{2mm}{\multirow{2}{*}{\rotatebox[origin=c]{90}{\textsc{en}}}}
	&\textsc{sup}-h      & 42.2 (4.7)           & 68.9 (9.7)            & 82.4 (3.0)            &  94.2 (0.8)\\
	&\textsc{m1+m2+bow}  & \textbf{63.5} (12.8) & \textbf{77.1} (2.8)   & \textbf{85.0} (1.5)   & -\\\midrule
	
	\parbox[t]{2mm}{\multirow{2}{*}{\rotatebox[origin=c]{90}{\textsc{de}}}}
	&\textsc{sup}-h      & 55.9 (9.9)           & 63.5 (10.2)           & 81.5 (6.5)            & 95.0 (0.3)\\
	&\textsc{m1+m2+bow}  & \textbf{63.2} (11.5) & \textbf{70.9} (6.3)   & \textbf{87.5} (2.6)   & -\\\midrule
	
	\parbox[t]{2mm}{\multirow{2}{*}{\rotatebox[origin=c]{90}{\textsc{fr}}}}
	&\textsc{sup}-h      & 38.6 (3.3)           & 55.9 (11.4)           & 78.5 (3.0)            & 93.5 (0.7)\\
	&\textsc{m1+m2+bow}  & \textbf{42.4} (4.6)  & \textbf{66.4} (9.1)   & \textbf{81.1} (2.7)   & -\\\midrule
	
	\parbox[t]{2mm}{\multirow{2}{*}{\rotatebox[origin=c]{90}{\textsc{ru}}}}
	&\textsc{sup}-h      & 49.4 (6.0)           & 53.8 (2.6)            & 68.2 (5.2)            & 87.2 (0.4)\\
	&\textsc{m1+m2+bow}  & \textbf{51.5} (6.0)  & \textbf{61.5} (4.6)   & \textbf{72.6} (2.3)   & -\\\midrule
	
	\parbox[t]{2mm}{\multirow{2}{*}{\rotatebox[origin=c]{90}{\textsc{zh}}}}
	&\textsc{sup}-h      & 63.4 (12.5)          & 70.7 (6.5)            & 81.2 (3.9)            & 91.1 (0.1)\\
	&\textsc{m1+m2+bow}  & \textbf{65.1} (11.1) & \textbf{77.1} (2.4)   & \textbf{81.4} (3.8)   & -\\
	  \bottomrule
	\end{tabularx}
    \vspace{-3mm}
     \caption{mBERT Encoder: MLDoc average test accuracy for both SUP-h and M1+M2+BOW models. The variance is in the bracket after the mean score. The first row denotes the number of labelled instances. The best results are in bold.} \label{app:tb:mbertmono}
     \vspace{-5mm}
\end{table*}

\begin{table*}[t]
\setlength{\tabcolsep}{2pt}
\scalebox{0.71}{
\begin{tabular}{p{0.3cm} | p{10cm} | p{10cm } }
\hline
\multicolumn{1}{c|}{\bf Class}& \multicolumn{1}{c|}{\bf M1+M2+BOW} &  \multicolumn{1}{c}{\bf M1+M2+GRU}\\
\hline
\multirow{2}{*}{\bf C}
&1: UNK, industry, credibility, agreement, ticket, co, decision, concept, ltd, people, sale, government, market, president, designations, minister, firm, plans, partner, deal
& 1: the bank said it lump of the united ... the new girls ltd said the concept ... the new extraordinary and the concept ... said the statement ...
\\
&2: UNK, ticket, year, shares, days, results, age, net, demand, securities, period, stock, concept, construction, bank, programme, procedure, statement, value, commission
& 2: the bank of organisation said on thursday that it had revoked by the first girls ...  first year to ...
\\\hline
\multirow{2}{*}{\bf E}
&1:
UNK, finance, market, loophole, budget, surprise, bank, basis, issue, government, system, exchanges, committee, municipal, world, securities, holding, net, confidence, minister
&1:
the international basic fund said on acknowledged that it said on publish to vote on publish to a bank said on publish ... 
\\
&2:
UNK, ticket, city, escalation, finance, bank, budget, concept, revenue, net, price, sale, trade, tax, prices, markets, series, rate, fund, pack 
&2:
the bank of submitting on publish florence said on acknowledged that ... it said on publish that ... to the new coherent said on acknowledged to bumping the bank said the bank ... 
\\\hline
\multirow{2}{*}{\bf G}
&1: UNK, government, state, minister, delay, pension, work, president, plans, summit, ticket, people, procedure, conference, ambassador, country, talks, opposition, nations, house 
& 1: the president remarkable said on thursday it surprise of ethnocide arrival the infidels of the islamic of the waterway the bank was ...
\\
& 2: UNK, state, president, war, police, office, authorities, problem, information, result, country, rights, committee, city, people, biodiversity, justice, health, securities, issue
&2: the summit in the authors and a virtual geological and the first time of the first party of the first time of ...
\\
\hline
\multirow{2}{*}{\bf M}
&1:
UNK, ticket, phase, market, government, minister, markets, banks, bank, budget, floor, points, rate, traders, procedure, strength, economy, finance, prices, loophole
&1:
the database distinctions the market closed sharply entire on thursday on acknowledged ...
\\
&2:
UNK, markets, market, stock, loophole, points, trade, shares, ticket, corporate, speaker, issues, fund, bank, group, exchanges, results, anticipation, companies, surprise
&2:
the following of the the the ries and not have embargo costs unveiling on publish pleading a impact of the japanese ... market and a bank was to be of the bank ...
\\\hline
\end{tabular}
}
\caption{Generated samples from M1+M2+GRU (BOW) for class C (\textit{Corporate/Industrial}), E (\textit{Economics}), G (\textit{Government/Social}), and M (\textit{Markets}). We randomly sample $\mathbf{z}_2$ from the prior while varying $y$.}
\label{app:tb:ucgen}
\end{table*}

\begin{table*}[t]
\setlength{\tabcolsep}{2pt}
\scalebox{0.71}{
\begin{tabular}{p{0.3cm} | p{10.4cm} | p{10.4cm }}
\hline
\multicolumn{3}{p{22.5cm}}{1: Fiat shares lost nearly two percent on Wednesday, slipping below the psychologically important 4,000 lire level in thin trading on a generally easier Milan Bourse, traders said. "The stock has gradually lost ground but without any major sell orders. At the moment there just isn't any interest in Fiat," one trader said. At 1439 GMT, Fiat was quoted 1.99 percent off at 3,980 lire, after touching a day's low of 3,970 lire, in volume of just under four million shares. The all-share Mibtel index posted a 0.47 percent fall. -- Milan newsroom +392 66129589 (E)}\\
\cdashline{1-3}
\multicolumn{3}{p{22.5cm}}{1: fiat shares lost nearly two percent on UNK slipping below the psychologically important UNK lire level in thin trading on a generally easier milan UNK traders UNK UNK stock has gradually lost ground but without any major sell UNK at the moment there just UNK any interest in UNK one trader UNK at UNK UNK fiat was quoted UNK percent off at UNK UNK after touching a UNK low of UNK UNK in volume of just under four million UNK the UNK UNK index posted a UNK percent UNK UNK milan UNK UNK UNK}\\
\hline
\multicolumn{3}{p{22.5cm}}{2: The top prosecutor of Honduras said on Wednesday that his country is a haven for money laundering. "In Honduras it's easy to launder money, the system allows it," Edmundo Orellana told reporters. "It's permitted because there is no law in Honduras that obligates a Honduran to explain the origin of his wealth." Honduran authorities estimate that $\$$300 million in illegal drug profits is laundered through the country each year. Money laundering is not classified as an offence in Honduras, although legislators have been working on a bill to outlaw it since last year. (G)}\\
\cdashline{1-3}
\multicolumn{3}{p{22.5cm}}{2: the top prosecutor of honduras said on wednesday that his country is a haven for money UNK UNK honduras UNK easy to launder UNK the system allows UNK UNK UNK told UNK UNK permitted because there is no law in honduras that UNK a honduran to explain the origin of his UNK honduran authorities estimate that UNK million in illegal drug profits is laundered through the country each UNK money laundering is not classified as an offence in UNK although legislators have been working on a bill to outlaw it since last UNK}\\
\hline
\multicolumn{1}{c|}{\bf Class}& \multicolumn{1}{c|}{\bf M1+M2+BOW} &  \multicolumn{1}{c}{\bf M1+M2+GRU}\\
\hline
\multirow{2}{*}{\bf C}
&1:
UNK, ticket, profit, concept, net, market, escalation, share, results, shares, delay, group, revision, profits, period, misery, statement, bank, key, procedure 
&1:
the bank said on fourthly it has inject requirement of the first group of ...
\\
&2:
UNK, concept, ticket, group, market, shares, delay, president, stock, companies, bank, statement, government, stake, price, co, state, girls, meeting, ltd 
&2:
the bank of organisation said on acknowledged that it had a meeting ...
\\\hline
\multirow{2}{*}{\bf E}
&1:
UNK, ticket, escalation, inflation, key, revision, delay, period, floor, consumer, bank, contexts, result, instance, show, market, level, government, gross, price 
&1:
the bank of submitting on publish florence said on acknowledged that it said on publish that ... the new coherent ... to the bank ... 
\\
&2:
UNK, ticket, bank, government, finance, market, state, budget, tax, minister, rate, delay, debt, issue, trade, investment, surprise, policy, sale, procedure
&2:
the international basic fund said on acknowledged that it said on publish ... to vote on acknowledged to a bank ... 
\\\hline
\multirow{2}{*}{\bf G}
&1:
UNK, world, ticket, policies, time, surprise, procedure, demand, campaigns, group, team, president, match, communities, place, minister, bank, government, number, relief
&1:
the ana police said acknowledged it had a tackling ... 
\\
&2:
UNK, president, government, people, state, minister, pension, police, designations, meeting, talks, opposition, leaders, country, security, result, statement, authorities, peace, summit
&2:
the president remarkable said on thursday that it surprise of ethnocide arrival infidels of her wines of her recall and the white house of ...
\\\hline
\multirow{2}{*}{\bf M}
&1:
UNK, shares, ticket, contexts, touch, market, stock, points, escalation, share, traders, phase, immigrants, procedure, price, pledges, revision, agriculture, group , level 
&1:
the bank of the settlement following the following vocational meda of the deal was delay ... and the market ...
\\
&2:
UNK, market, ticket, bank, traders, anticipation, delay, procedure, trade, prices, immigrants, rate, government, money, meda, escalation, demands, exchange, points, reallocation 
&2:
the bank of the settlement following the following vocational value of the relative gains of ...
\\\hline
\end{tabular}
}
\caption{Generated samples from M1+M2+GRU (BOW) by varying class label $y$. We take $\mathbf{z}_2$ from the input examples shown above. For each example, the first is the original document with the class label in the end, and the second is the real input to the system.}
\label{app:tb:cgen}
\end{table*}